%% file: weakly_supervised_house.tex
\documentclass{article} 
\usepackage{nips15submit_e,times}
\usepackage{hyperref}
\usepackage{url}
\usepackage{epsfig}
\usepackage{graphicx}
\usepackage{amsmath,amssymb} 
\usepackage{color}
\usepackage{comment}
\usepackage[ruled,linesnumbered]{algorithm2e}
\usepackage{wrapfig}

\nipsfinalcopy 
\title{Feedback Neural Network for Weakly Supervised Geo-Semantic Segmentation}

\author{
  Xianming Liu$^{1, 3}$, Amy Zhang$^2$, Tobias Tiecke$^1$, Andreas Gros$^2$, Thomas S. Huang$^3$ \\
  $^1$Facebook Connectivity Lab\\
  $^2$Facebook Core Data Science\\
  $^3$University of Illinois, Urbana-Champaign\\
  \texttt{\{xmliu, amyzhang, ttiecke, andreasg\}@fb.com} \\
  \texttt{t-huang1@illinois.edu}
}

\begin{document}

\maketitle

\begin{abstract}
Learning from weakly-supervised data is one of the main challenges in machine learning and computer vision, especially for tasks such as image semantic segmentation where labeling is extremely expensive and subjective.
In this paper, we propose a novel neural network architecture to perform weakly-supervised learning by suppressing irrelevant neuron activations. It localizes objects of interest by learning from image-level categorical labels in an end-to-end manner.
We apply this algorithm to a practical challenge of transforming satellite images into a map of settlements and individual buildings. Experimental results show that the proposed algorithm achieves superior performance and efficiency when compared with various baseline models.
\end{abstract}

\input{1_intro}
\input{2_related}
\input{3_weakly}
\input{4_impl}
\input{5_exp}
\input{6_conclusion}

\small
\bibliographystyle{unsrt}
\bibliography{egbib}

\end{document}

%% file: 1_intro.tex
\section{Introduction}
Answering ``what'' and ``where'' in images has been one of main challenges in computer vision.
With the development of Convolutional Neural Networks (CNN) in recent years, the accuracy of image recognition has been boosted significantly and even outperforms human beings \cite{russakovsky15_imagen_large_scale_visual_recog_chall, szegedy15_going}.
However, although tremendous efforts have been spent on object localization and segmentation,
it is still far away from competing with human performance.

From a machine learning perspective, semantic segmentation / object localization are limited heavily by the data employed during training. Compared with classification, the target is to train machine learning models with much finer supervisions, i.e., object bounding boxes or pixel-wise semantic labels, which is extremely time costly in acquisition \cite{lin14_micros_coco}.
Moreover, fine scale supervision is more subjective \cite{oquab15_is} and easily introduces errors.
As shown in the BSD image segmentation dataset\cite{martinil}, human labeling could only achieve a F1-Score of $0.79$ on a 500-image segmentation dataset.

In this paper, we consider a new strategy to perform image semantic segmentation / object localization in a \emph{weakly-supervised} manner. By weakly-supervised, our target is to let neural networks learn `what' and `where' simultaneously: we train a neural network with image level categorical labels only and predict pixel-level semantic labels.
In analogy to human cognition, who are not learning to recognize visual objects from pixel level labeled training data,
the motivation of the proposed work is to train neural networks to learn from large scale of data and be able to capture the common attributes of semantic concepts.

The implementation of the weakly-supervised neural network is inspired by several evidences from cognitive research, and is abstracted into a computational model.
As studied in visual cognition, individual visual cortex neuron/neuron populations are responding to different objects or components \cite{dicarlo12_how_does_brain_solve_visual_objec_recog,kruger13_deep_hierar_primat_visual_cortex,stansbury13_natur_scene_statis_accoun_repres};
Though there are no evidences showing that neural networks are mimicking human brains, visualization of individual neurons indicates that CNNs try to train their  kernels to capture objects and semantics at different levels \cite{zeiler14_visual_under_convol_networ, cao15_look_think_twice}.
Furthermore, the key for object recognition and localization from a cluttered scene for humans is visual attention, which is driven by \emph{Biased Competition Theory} \cite{desimone98_visual_atten_mediat_by_biased, beck09_top_down_bottom_up_mechan}: unrelated signals will be suppressed during a feedback loop, and the selectivity of neuron activations functions to make foreground ``pixels'' salient \cite{desimone95_neural_mechan_selec_visual_atten}.
Studies in computer vision and machine learning have utilized this mechanism for various tasks in recent years, such as Part Based Models \cite{felzenszwalb10_objec_detec_with_discr_train} and Feedback Neural Networks \cite{cao15_look_think_twice}.
Moreover, different from current vanilla CNN computational models, using feedback has shown its effectiveness in object recognition and detection in both cognitive research \cite{Cichy2014Resolving} and machine learning \cite{hu15:_bottom, cao15_look_think_twice}.
In this paper, we formulate the neuron suppression (as selectivity) and feedback as an efficient optimization process in CNNs, by introducing feedback layers and a feedback loop besides the traditional forward-backward procedure. It successfully predicts pixel level semantic labels with accurate boundaries in an end-to-end framework.

The proposed \emph{Weakly-Supervised Learning} scheme provides the solution to a broad range of real-world problems where large amounts of finely labeled training samples are not available. As an example, in this paper we show the application of creating country-wide geographical maps based on satellite imagery, by finding the ``footprints'' of man-made architectures.
It takes binary (``There is a building / no building in this image'') image-level labeled data for training, and gives pixel level semantic segmentation as output.
This task is extremely challenging since:
\\\emph{Supervision}: there is no finely labeled data in training in this approach;
\\\emph{Quality}: all training and testing data are of low resolution, and potentially suffer from noise, large intra-class variances, occlusion, and high contrast changes;

%
Experimental results show that the proposed weakly-supervised network achieves satisfying performance, and is much superior and more stable than all other weakly-supervised neural networks.

%% file: 2_related.tex
\section{Related Work}
\subsection{Image Semantic Segmentation}
Image semantic segmentation aims at assigning each pixel a semantic label. Most algorithms incorporate both global and local information during inference: global information is used to decide ``what'' and local information gives ``where'' \cite{long15_fully}.
Recently, with the strong discriminative ability of deep neural networks, several architectures have been proposed to insert local information into the end-to-end classification procedure and produce pixel-by-pixel semantic segmentation results. Long \emph{et. al.} propose a Fully Convolutional Network (FCN) \cite{long15_fully} to extend neural networks to spatially dense predictions on pixel level.
Similarly, Badrinarayanan \emph{et. al.} adopt an encoder-decoder style network to map the classification results back to pixel level labels \cite{badrinarayanan15:_segnet}.
To increase the localization accuracy, Ronneberger \emph{et.al.} design a symmetric network architecture to take deeper layers for classification and shallower layers for localization \cite{ronneberger15_u_net}.
Other more recent methods incorporate a Conditional Random Field to further improve the localization accuracy and reduce outliers \cite{zheng15_condit_random_field_recur_neural_networ,chen14:_seman_image_segmen_deep_convol}.

However, all these algorithms require a large amount of pixel-level finely labeled training data, which is time consuming to acquire and subjective to use. To overcome this limitation, weakly-supervised learning algorithms are proposed to bridge the gap between training data and the target.
\subsection{Weakly-Supervised Learning}
Learning from weakly-supervised data emerged recently as abundant weakly labeled data became available to computer vision and machine learning research.
All approaches can be mainly divided into two categories, i.e., Multiple Instance Learning (MIL), and Self-Training (bootstrapping).

\noindent
\textbf{MIL}:
The most straightforward formulation of weakly-supervised learning is via MIL. It assumes the weakly-supervised label as a positive instance of the ``true hypothesis''.
Given the image level categorical labels, which indicate the presence or absence of the target class in the training sample, it tries to find the common signal present in positive images but absent from negative ones \cite{song14}. Either supervised detectors \cite{siva11_weakl} or unsupervised learning of mid-level visual elements \cite{endres13_learn_collec_part_model_objec_recog} are frequently used to discover semantic-meaningful instances.

More recent approaches utilize the region proposal methods, such as Selective Search \cite{uijlings13_selec_searc_objec_recog}. Selective search has been proven to be efficient in object detection, and could give meaningful region proposals, especially working together with CNNs.
In \cite{song14}, Song et. al. propose to automatically discover positive region proposals using a MIL support vector machine, based on neural network features. And in \cite{liweakly2016}, Li et. al. propose a two steps procedure to sequentially optimize neural network (classification adaptation) and the detector (detection adaptation).

MIL has shown its efficiency on the task of object localization. However, it is dependent on the quality of instances: object detector, middle-level visual elements or low-level region proposals.
It limits these algorithms when applied to much finer tasks such as semantic segmentation on the pixel-level.
In addition, MIL relies on its initialization, especially initializations of detectors. Therefore, MIL formulation is hard to generalize to large scale dataset with high intra class variances.

\noindent
\textbf{Bootstrapping(Self-Training)}:
For Bootstrapping (Self-Training), a model is trained on the reference space (e.g., on image classification), and predictions on the target domain (e.g., semantic segmentation) are used to train the target together with suitable constraints.
It works well with Deep Neural Networks: a classification neural network is trained on image level data, which will produce pixel wise predictions.
Various constraints have been developed to generate reasonable pixel wise segmentations.
For example, in \cite{papandreou15_weakl_semi_super_learn_deep}, George Papandreou \emph{et. al.} propose to use spatial local consistency as the constraint and develop a CRF layer to propagate the pixel level semantic labels.
In \cite{pathak15_const_convol_neural_networ_weakl_super_segmen}, Pathak \emph{et. al}. use the constraint that the foreground pixels should have larger response than the background ones, to guide the backpropagation.
All these methods report good performance on semantic segmentation and can easily generalize to large scale of data.
Despite the positives, these approaches still need a certain amount of fully supervised data to achieve reasonable performance. Moreover, they are sensitive to foreground size and usually perform badly on small objects.

To this end, we propose a novel weakly-supervised semantic segmentation routine by applying an extra Feedback loop in addition to the feedforward procedure in the neural network. During the feedback stage, the network is designed to suppress all irrelevant neuron activations by performing a relaxed approximation of a sparse optimization.
Compared with the feedback network proposed in \cite{cao15_look_think_twice}, our solution is much more efficient while produces high quality inference results: instead of solving the optimization iteratively, the algorithm proposed in this paper derive satisfying results in one step.
Moreover, different from \cite{cao15_look_think_twice}, where only the gradient map could be generated for the purpose of object localization, our algorithm can predict dense pixel-level probabilistic maps.

%% file: 3_weakly.tex
\section{Weakly-Supervision via Feedback in Neural Network}

Given training data and supervision $\{x_i, y_i\}_{i=0}^L$, our neural network tries to minimize the loss function
\begin{equation}
f_w = \arg\min{\sum_i \frac{1}{2} \|y_i - f_w(x_i)\|^2 + \lambda \|w\|_2},
\end{equation}
\noindent
where $f_w$ is the transformation from input $x$ to output $\hat{y}$ parameterized with $w$.

Convolutional Neural Networks (CNNs) perform well at characterizing visual elements from low-level to higher semantic levels, from basic visual patterns (e.g., Gabor Filters) to contours, object parts and whole objects \cite{zeiler14_visual_under_convol_networ,mahendran15_under}.
Inspired by the \emph{Biased Competition Theory}, in which the human visual cortex only selects the relevant stimuli and suppresses irrelevant ones,
a feasible solution for weakly-supervised semantic segmentation could work by suppressing activations of neurons irrelevant to the current target.

Considering a particular layer of neural network $l$, with input $x^l$ and target output $y^l$, we could interpret this as a computational model by optimizing the target function of:
\begin{equation}
\min \frac{1}{2} \|y^l - f_w(x^l)\|^2 + \gamma \|x^l\|_1,
\label{eqn:target}
\end{equation}
\noindent
which tries to optimize the target output $f_w(x^l)$ and control the activations of neurons $x^l$.
Note that we omit the $L2$ regularization term $\|w\|_2$ since it could be easily implemented using weight decay in Stochastic Gradient Descent.

To solve this optimization, we utilize the solution in \cite{lee2006efficient}, which tries to optimize two subproblems, $L_1$-Regularized least squares problem and $L_2$ constrained least squares problem:
\begin{itemize}
\item If $\frac{\partial \|y^l - f_w(x^l)\|^2}{\partial x_i^l} > \gamma$, then neuron $x_i^l$ should be positively activated,
\item If $\frac{\partial \|y^l - f_w(x^l)\|^2}{\partial x_i^l} < -\gamma$, then neuron $x_i^l$ should be negatively activated,
\item Otherwise, neuron $x^l_i$ should be deactivated.
\end{itemize}
And in the application of semantic segmentation, we further require the neuron activation to be positively related with the target $y$.

This gives an approximate solution to the $L_1$ subproblem shown in Equation \ref{eqn:target}. To solve the $L_2$ constrained subproblem, a standard back-propagation in Stochastic Gradient Descent is performed.
For multi-layered neural networks, the optimization is greedily performed in a layer-wise fashion. And similar to \cite{cao15_look_think_twice}, we train networks and test examples in a feedback procedure in addition to the standard feedforward-backward routine.
However, compared with \cite{cao15_look_think_twice}, in which multiple iterations of feedback are used during testing, the proposed solution does not run iteratively, and hence the efficiency is largely improved.
We summarize the procedure in Algorithm \ref{alg:weakly-supervised}.

\begin{algorithm}[htb]
\KwIn{Sample $\{x_i, y_i\}$, and neural network initialization $f_w$}
\emph{Forward}: Inferring estimation $\hat{y}_i = f_w(x_i)$;

\emph{Feedback}: Estimating Neuron Activations $\mathbb{\theta}$

\For{each layer $l$}
{
    \uIf{$\frac{\partial \|y^l - f_w(x^l)\|^2}{\partial x_i^l} > \gamma$}
    {
        Activate neuron $x_i^l$: Set $\theta_i^l = 1$;
    }
    \uElse
    {
        Suppress neuron $x_i^l$: Set $\theta_i^l = 0$;
    }
}

\emph{Backward}: Perform backpropagation and update parameter using SGD.

\caption{Weakly-Supervised Learning via Feedback Network}
\label{alg:weakly-supervised}
\end{algorithm}

\subsection{Revisiting the Feedback Routine}
For the input $x^l$ of a particular layer $l$, the final estimation of neural network is denoted as
\begin{align}
\hat{y} = f_w^l(x^l) \cong T^l \cdot x^l + O(x^2)
\end{align}
\noindent
using first order Taylor Expansion.
Especially, for a determined input, all nonlinear neurons in transformation $f_w^l$ are deterministic, and $f_w^l$ becomes a linear transformation related with input $x$.
We denote the linear transformation as $T^l(x)$.
Taking the derivative on both sides gives
\begin{equation}
\frac{\partial \hat{y}}{\partial x^l} = T^l(x)
\end{equation}
\noindent
and similarly
\begin{equation}
\frac{\partial \|y - \hat{y}\|^2}{\partial x^l} \propto T^l(x).
\end{equation}

Compared with Algorithm \ref{alg:weakly-supervised}, we could explain the proposed algorithm in a straightforward way: The optimization in feedback procedure tries to activate those neurons $x_i$ with $T^l_i(x) > \gamma$, which means it has certain positive contributions to the final target and suppresses those with negative contributions.
It also gives a simple method for testing: In order to locate the object(s) related with target concept $y_i$, set the gradient vector $\delta = y - \hat{y} = [0, 0, ..., 1, 0, ...]$, where only the $i$-th dimension $\delta_i$ is non-zero; and run the feedback procedure in Algorithm \ref{alg:weakly-supervised} to get the corresponding neuron activations.

%% file: 4_impl.tex
\section{Implementation Details}
\subsection{Implementation}

\begin{wrapfigure}{r}{0.25\textwidth} 
  \centering
  \includegraphics[width=0.25\textwidth]{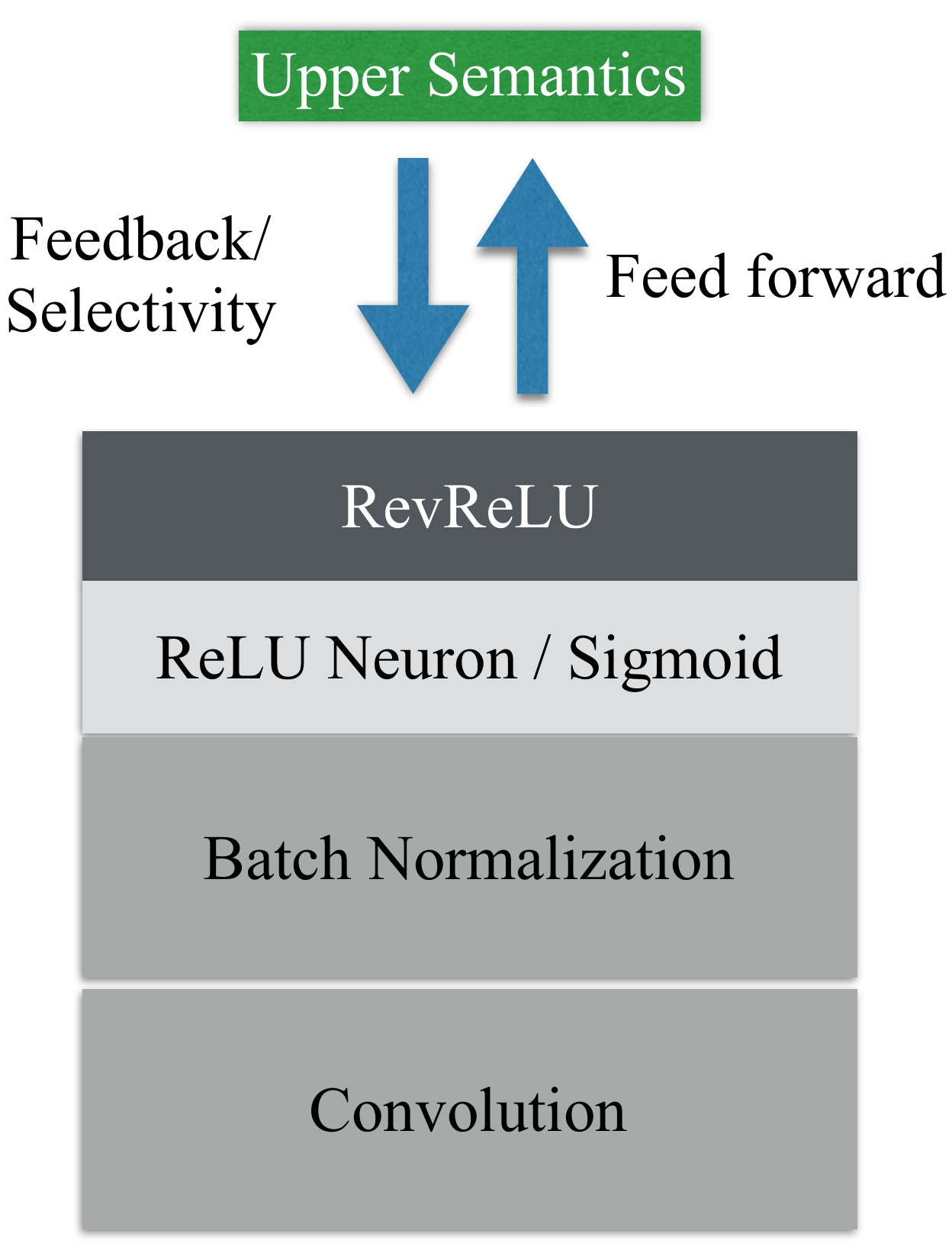}
  \caption{The design of Feedback Unit.}
  \label{fig:revrelu}
  \vspace{-10pt}
\end{wrapfigure}

To implement the proposed network architecture, we introduce the implementation of \emph{Feedback Layers}, which is designed to perform the neuron activation selection.
Feedback layers behave similarly to ReLU layers, except it allows only neurons with positive gradients being activated, \emph{i.e.}, $\frac{\partial \|y - \hat{y}\|^2}{\partial x^l} > \gamma$, while ReLU only allows those with positive outputs being activated.
To further improve the efficiency, we relax $\gamma$ to be $0$, and name the layer  ``RevReLU'' in analogy to the ReLU layer. RevReLU layer is stacked upon ReLU layer, as shown in Figure \ref{fig:revrelu}, to complete the feedback optimization in Algorithm \ref{alg:weakly-supervised}.
In practice, we build the ``Feedback Unit'' composed of Convolution layer + Batch Normalization Layer + ReLU layer + RevReLU layer.

\begin{figure}
\centering
  \includegraphics[width=0.8\linewidth]{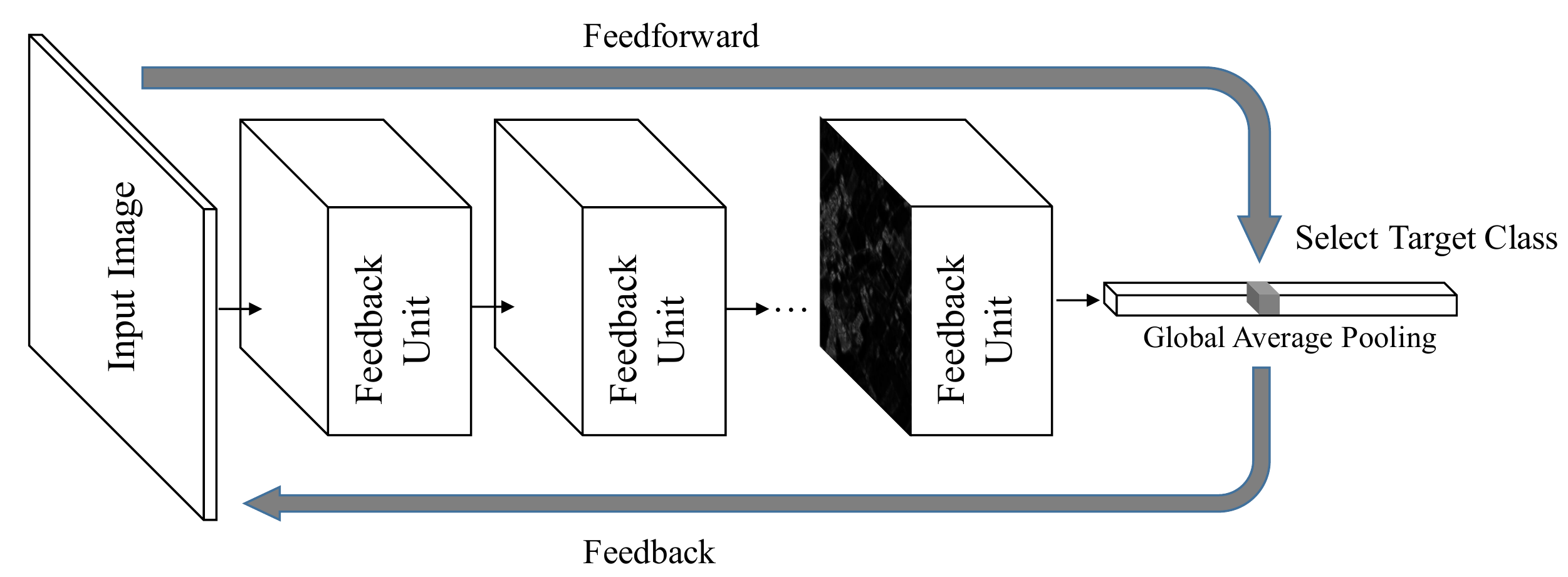}
  \caption{Weakly-Supervised Neural Network architecture. Global Average Pooling is used to map the pixel wise probability to classification result, and the last convolution layer is used to derive semantic segmentation result.}
  \label{fig:network_arc}
  \vspace{-10pt}
\end{figure}

The weakly-supervised network is built by stacking multiple \emph{feedback units}, as shown in Figure \ref{fig:network_arc}.
The number of channels in the last convolution layer equals to the number of classes to output,
and the output probability is derived by applying global average pooling.
We also use the neuron activations of the last convolution layer as the semantic segmentation probabilistic maps.

\subsection{Training and Inference}

For training, we employ standard vanilla convolutional neural network training strategy, which takes a training sample with image level categorical labels, and trains a classification neural network.
Alternatively, it is also optional to take feedback into training, which is more time consuming (triple the training time) but results in better localization accuracy. To balance the performance and efficiency, we use vanilla style training (feedforward training) in all our experiments.

During the testing time, for input image $x$, the network first performs a forward pass and gets the estimation $\hat{y} = f_w(x)$.
As shown in Figure \ref{fig:network_arc}, the top $1$ (or top $k$) target neuron $j$ is selected as ``what is there in this image'',  and a feedback procedure is performed to find ``where'' by suppressing irrelevant neuron activations. After running the forward once more, we harvest the output of the last convolution layer, and use the channel $j$ as the dense localization probabilistic map for target $j$.
Note that this strategy also works for finding multiple objects of multiple classes.

\subsection{Miscellaneous Improvements}

There are mainly two difficulties in using the convolutional layer output directly to localize objects of interest:
\noindent
\\\textbf{Down Sampling of Spatial Resolution}: To fit large images into memory, it is necessary to shrink the convolution output size in convolutional neural networks. We employ a similar method to \cite{long15_fully}, using a deconvolution layer on top of the final convolution to up-sample the probability map into the same resolution as the input image.
\\\textbf{Location Shift}: As more convolutions are applied, the receptive field of each neuron increases and the localization accuracy of boundary pixels decreases. We simply compose the input image with the probability map together by using an element-wise product, which is similar to U-Net \cite{ronneberger15_u_net}.

Other tricks including using Sobel filter and watershed segmentation algorithm to perform morphological operations, and improve the final output.

%% file: 5_exp.tex
\section{Experiments}

\subsection{Problem Setup}

\setlength{\tabcolsep}{0.4pt}
\begin{figure}[htb]
\vspace{-10pt}
\begin{center}
\begin{tabular}{cccccccc}
\includegraphics[width=0.10\linewidth]{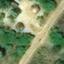} &
\includegraphics[width=0.10\linewidth]{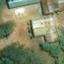} &
\includegraphics[width=0.10\linewidth]{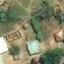} &
\includegraphics[width=0.10\linewidth]{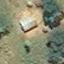} &
\includegraphics[width=0.10\linewidth]{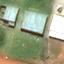} &
\includegraphics[width=0.10\linewidth]{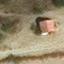} &
\includegraphics[width=0.10\linewidth]{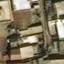} &
\includegraphics[width=0.10\linewidth]{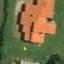} \\
\includegraphics[width=0.10\linewidth]{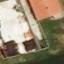} &
\includegraphics[width=0.10\linewidth]{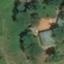} &
\includegraphics[width=0.10\linewidth]{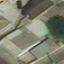} &
\includegraphics[width=0.10\linewidth]{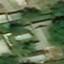} &
\includegraphics[width=0.10\linewidth]{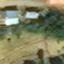} &
\includegraphics[width=0.10\linewidth]{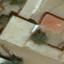} &
\includegraphics[width=0.10\linewidth]{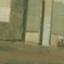} &
\includegraphics[width=0.10\linewidth]{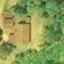}
\end{tabular}
\caption{Examples of satellite images (Digital Globe) of positive 64 by 64 patches, sampled from 4 different countries.}
\label{fig:patch_examples}
\end{center}
\vspace{-10pt}
\end{figure}

In this paper, we are testing our model on a practical real world problem: we apply the proposed weakly-supervised neural network on the semantic segmentation of satellite imagery data, and compare with several state-of-the-arts.
Our goal is to develop an automated mapping tool to identify human-made settlements as a proxy for where people occupy land and hence create a geographical map of the population distribution.

We do this by finding the ``footprints'' of buildings at a country-wide level from satellite images with special preference on rural areas, which is even more challenging that all targets are of tiny sizes and the foreground is highly sparse.
Both the training and testing data originates from 20 countries, is spread out across multiple continents and ranges from urban areas, to rural areas, and even to deserts / forests.
Labeling such imagery data on pixel level or bounding box level is extremely challenging, considering its quality and spatial resolution. We show a series of examples in Figure \ref{fig:patch_examples}, where a house usually occupies only tens of pixels. Labeling such data will also introduce unbounded noise and errors in supervision.
Another challenge is its volume: it takes around 20 TB of imagery per country, and its imbalanced distribution: typically over 95\% of the landmass does not contain a building. To get representative testing data, a large amount of labeled samples are required.

\noindent
\textbf{Training and Testing Data}
To this end, we adopt the weakly-supervised learning mechanism:
$64*64$ patches are sampled from all countries, with only binary image-level labels: $1$ means \emph{there is a house / are houses in this patch}, and $0$ for \emph{none}. The sampling is performed randomly for testing data and based on a weak classifier to obtain training data.
This results in a training data set of about $10$k images per country ($220$k total), with almost balanced ratio for both two classes.
Moreover, 8-fold augmentation is applied to generate more diverse training data. Finally, a training set containing $1.76$ million weakly-supervised data is harvested.

As for the testing data, we cherrypicked 8 satellite imagery tiles with spatial size 1 square kilometer (or equivalently 2048 pixels by 2048 pixels) from both rural areas and urban areas.
To get the groundtruth for evaluation purpose, each tile is labeled by at least two well-trained labelers and those with conflicts are eliminated from the evaluation set.
All algorithms are required to take those $2048 \times 2048$ images as input and provide pixel wise semantic segmentation as output.

\noindent
\textbf{Network Structure}
We design our network structure following the guidance in Figure \ref{fig:network_arc}, and using feedback unit in Figure \ref{fig:revrelu} as basic building blocks.
The network contains 7 feedback units in total, and uses one deconvolution layer to upsample the output back to the original image size.

\subsection{Performance Evaluation}
\noindent
\textbf{Evaluation Criteria}: We evaluate the semantic segmentation on the pixel-level, by viewing it as a pixel-wise classification. The optimal \emph{F-Score} is utilized as evaluation criteria, especially considering the extremely unbalanced foreground / background distribution.

\noindent
\textbf{Baseline methods}: There are few neural networks able to perform semantic segmentation by training from merely weakly-supervised data, as used in our experiments.
Without loss of generalization, we compare our proposed algorithm with two main-stream baselines trained using Neural Networks, U-Net \cite{ronneberger15_u_net}, and SegNet \cite{badrinarayanan15:_segnet}.
In order to make these baseline networks able to train on weakly-supervised data, we modify the net by adding a global average pooling layer on top of pixel-wise outputs to generate image level predictions.

\begin{wraptable}{r}{0.4\linewidth}
\vspace{-12pt}
\centering
\small
\caption{Comparison on AUC scores.}
\begin{tabular}{c|c}
\hline
                                       &F-Score \\ \hline
U-Net \cite{ronneberger15_u_net}       &0.260 \\ \hline
SegNet \cite{badrinarayanan15:_segnet} &0.199 \\ \hline
Proposed Net w/o Feedback              &0.407 \\ \hline
Proposed Net w. Feedback               &\textbf{0.414} \\ \hline
\end{tabular}
\label{tab:performance}
\end{wraptable}

\begin{figure}[htb]
\begin{center}
\begin{tabular}{cccc}
\textbf{Original Image} & \textbf{SegNet} & \textbf{U-Net} & \textbf{Feedback Net} \\
\includegraphics[width=0.24\linewidth]{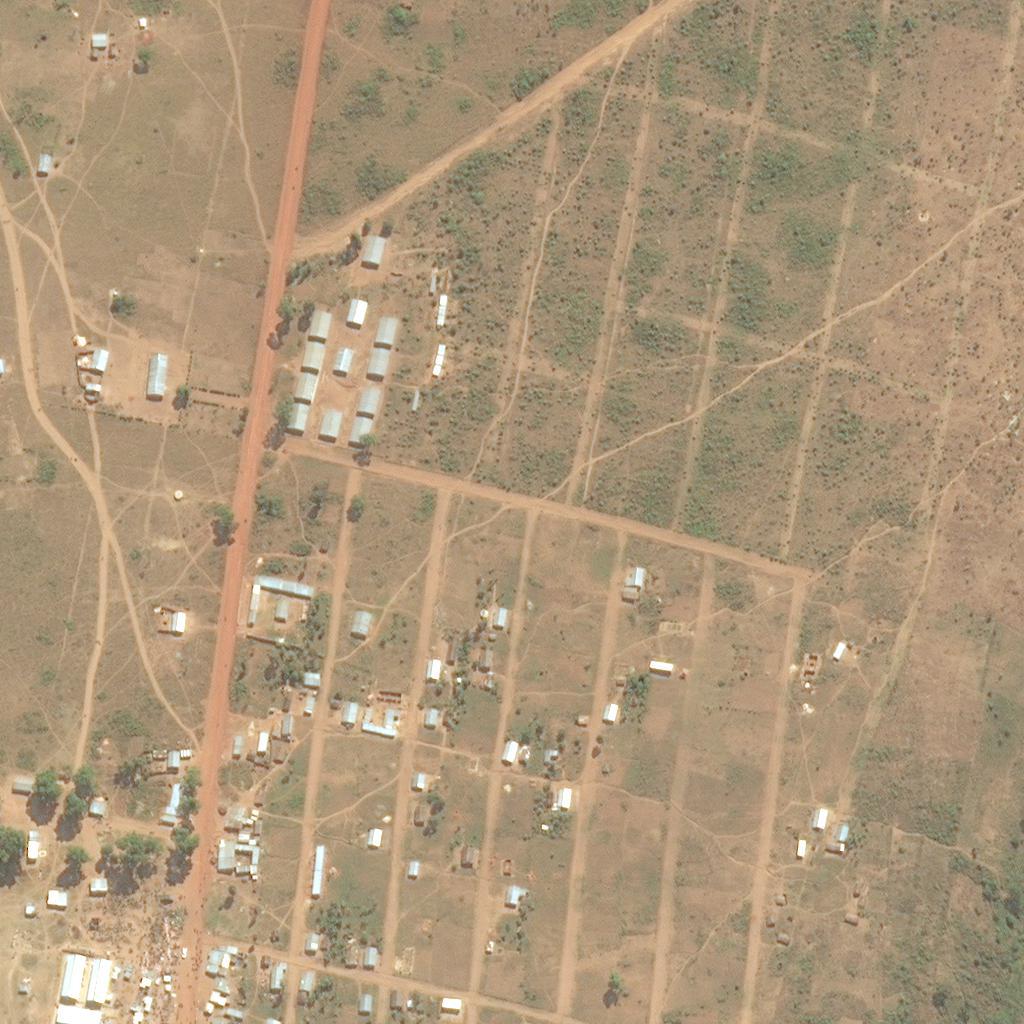} &
\includegraphics[width=0.24\linewidth]{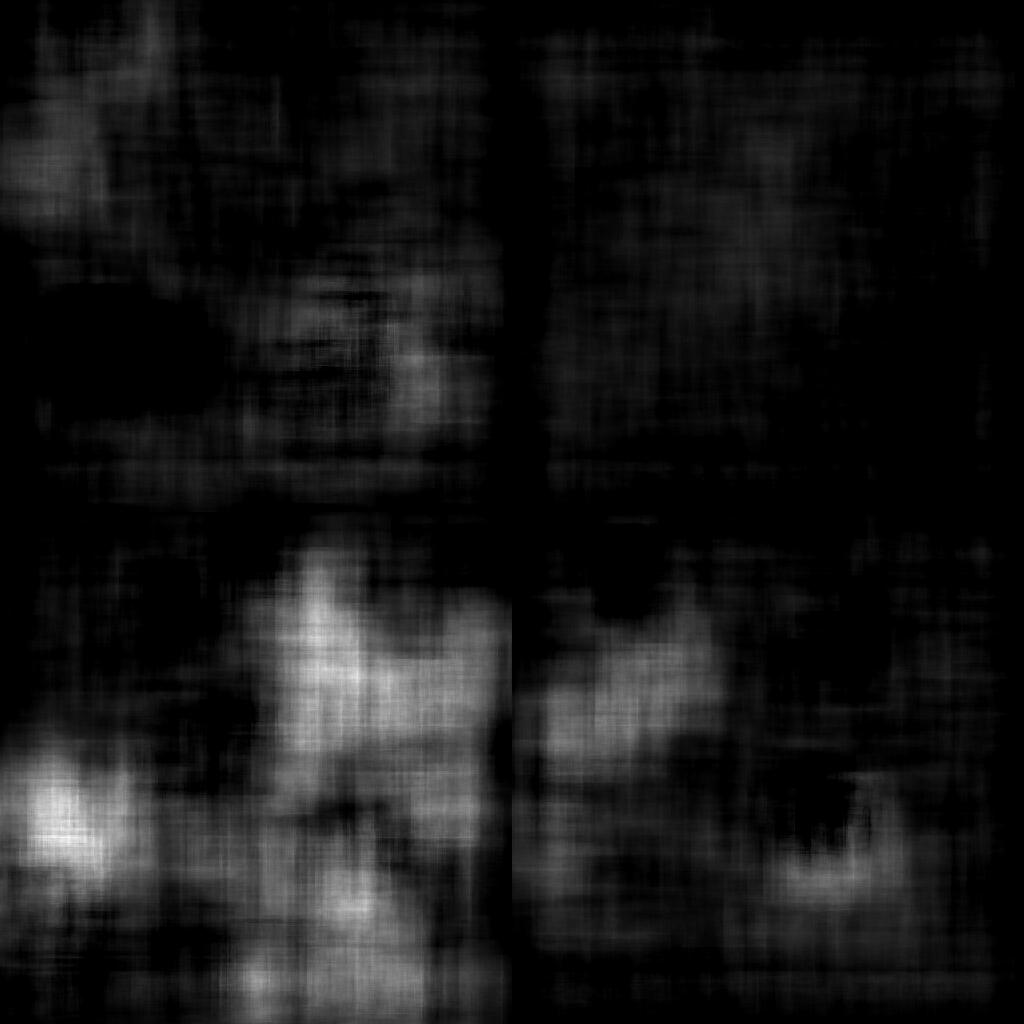} &
\includegraphics[width=0.24\linewidth]{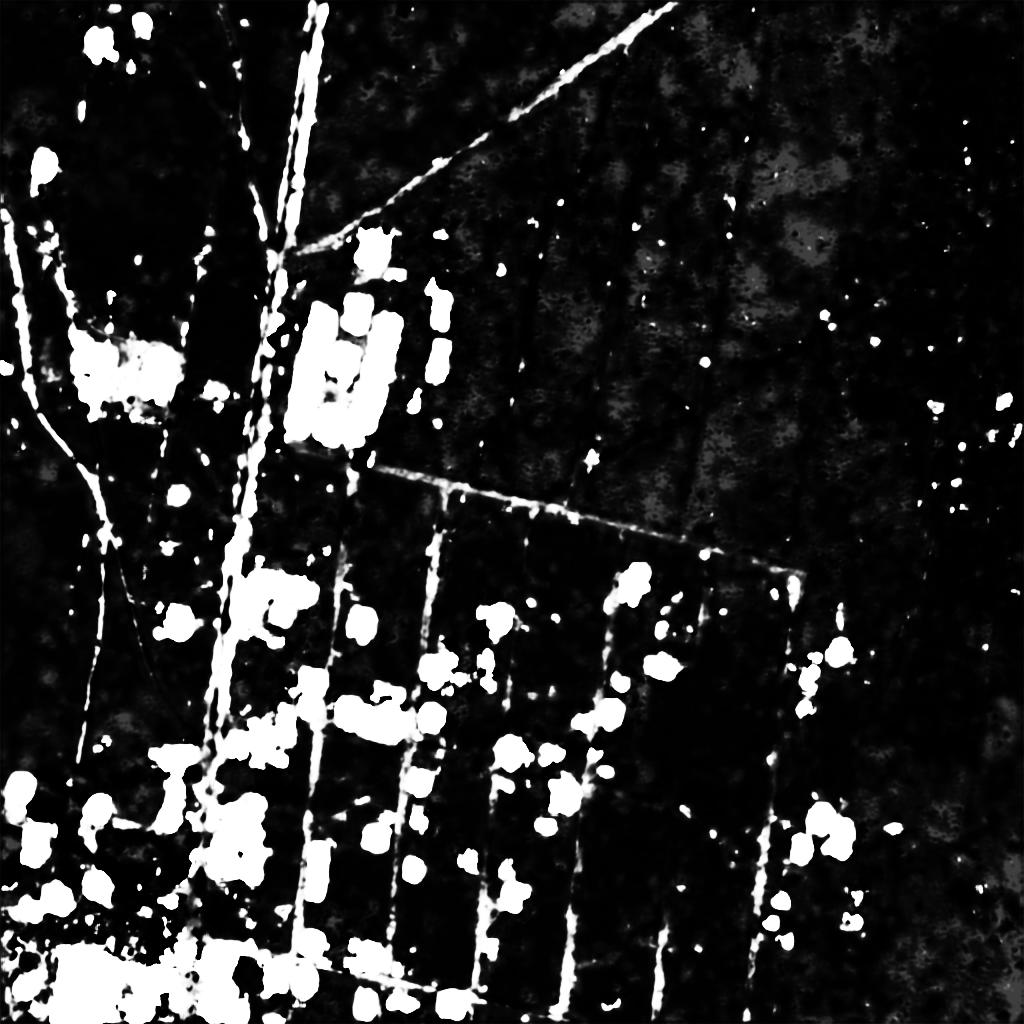} &
\includegraphics[width=0.24\linewidth]{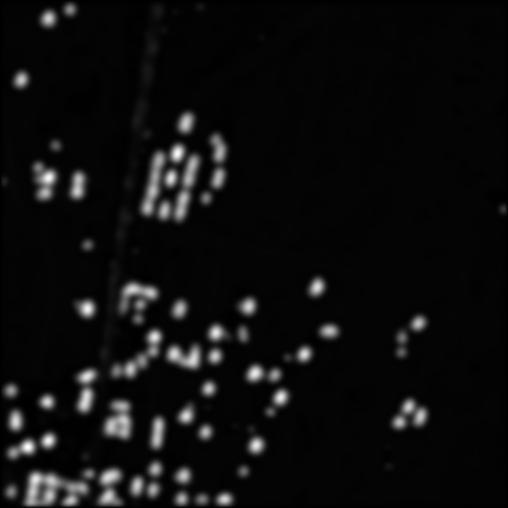} \\
\includegraphics[width=0.24\linewidth]{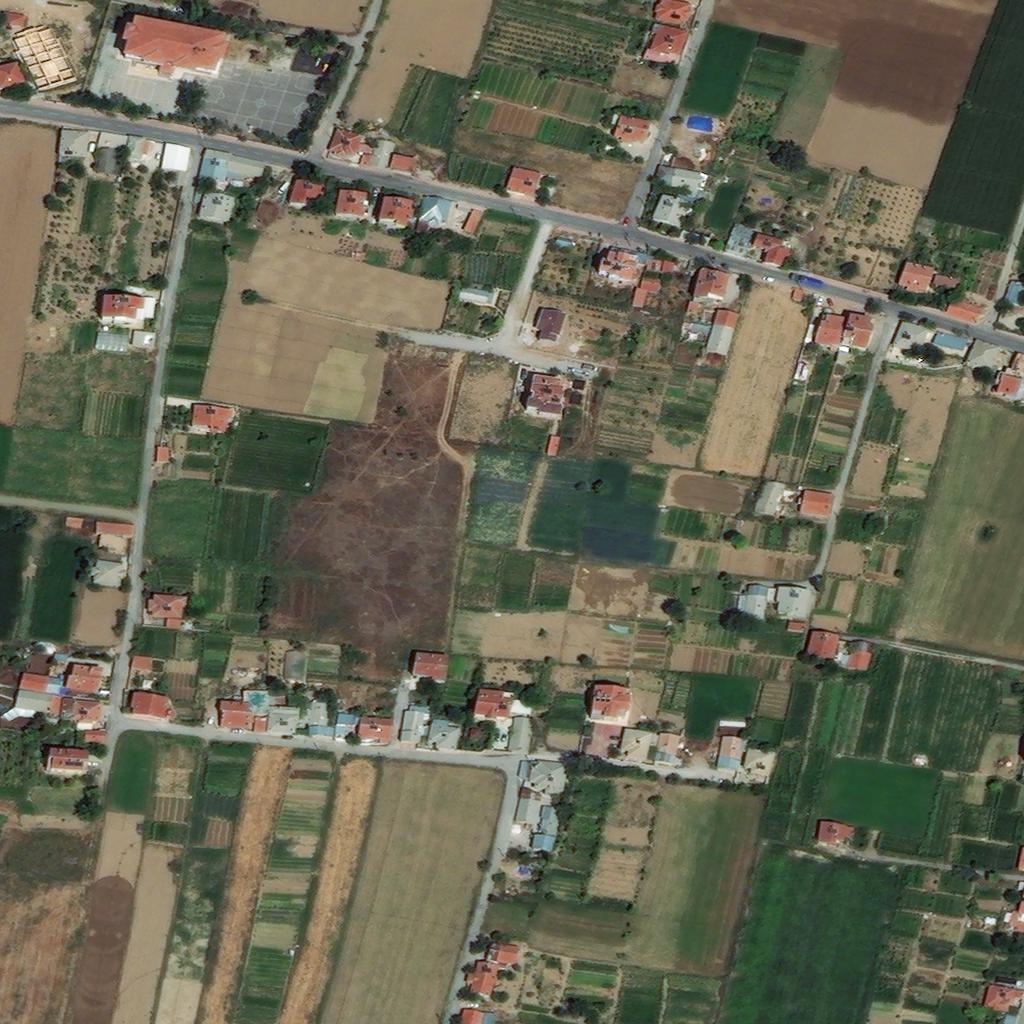} &
\includegraphics[width=0.24\linewidth]{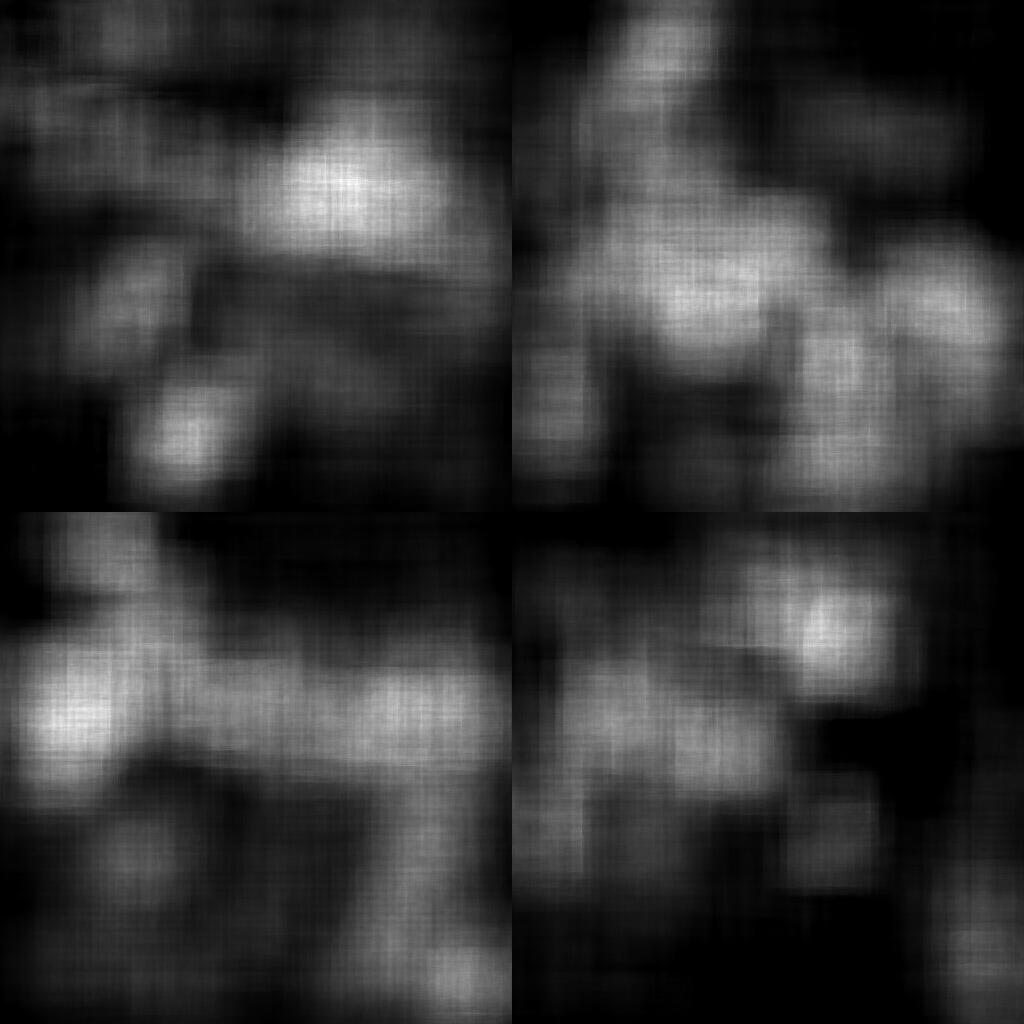} &
\includegraphics[width=0.24\linewidth]{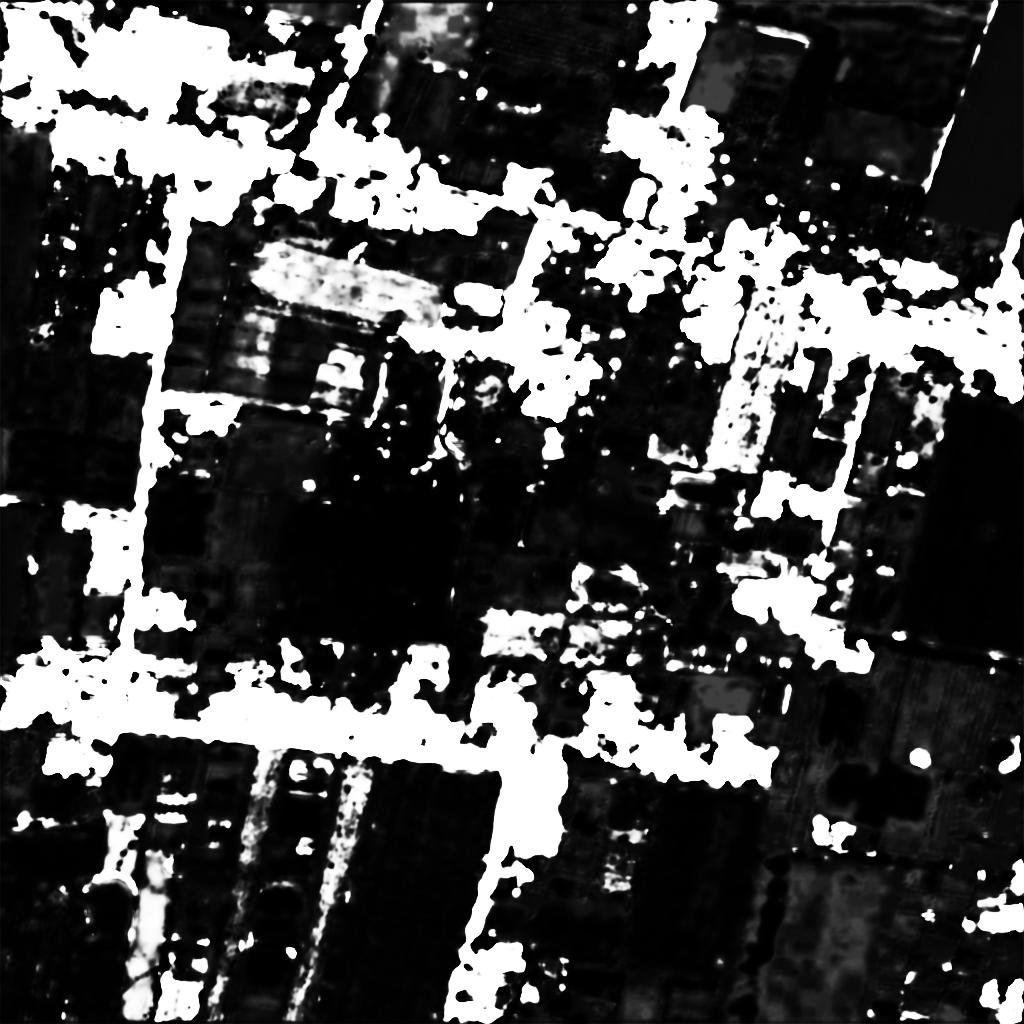} &
\includegraphics[width=0.24\linewidth]{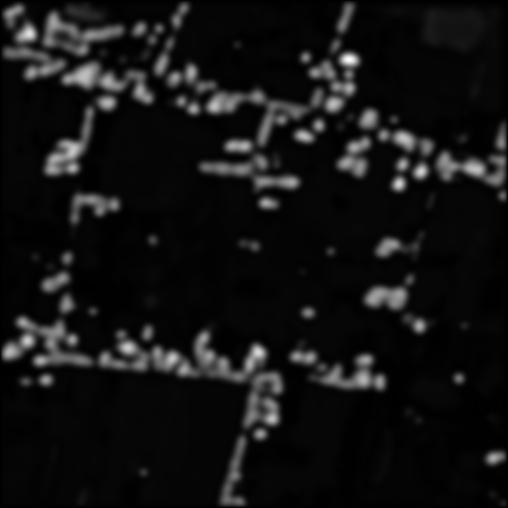} \\
\includegraphics[width=0.24\linewidth]{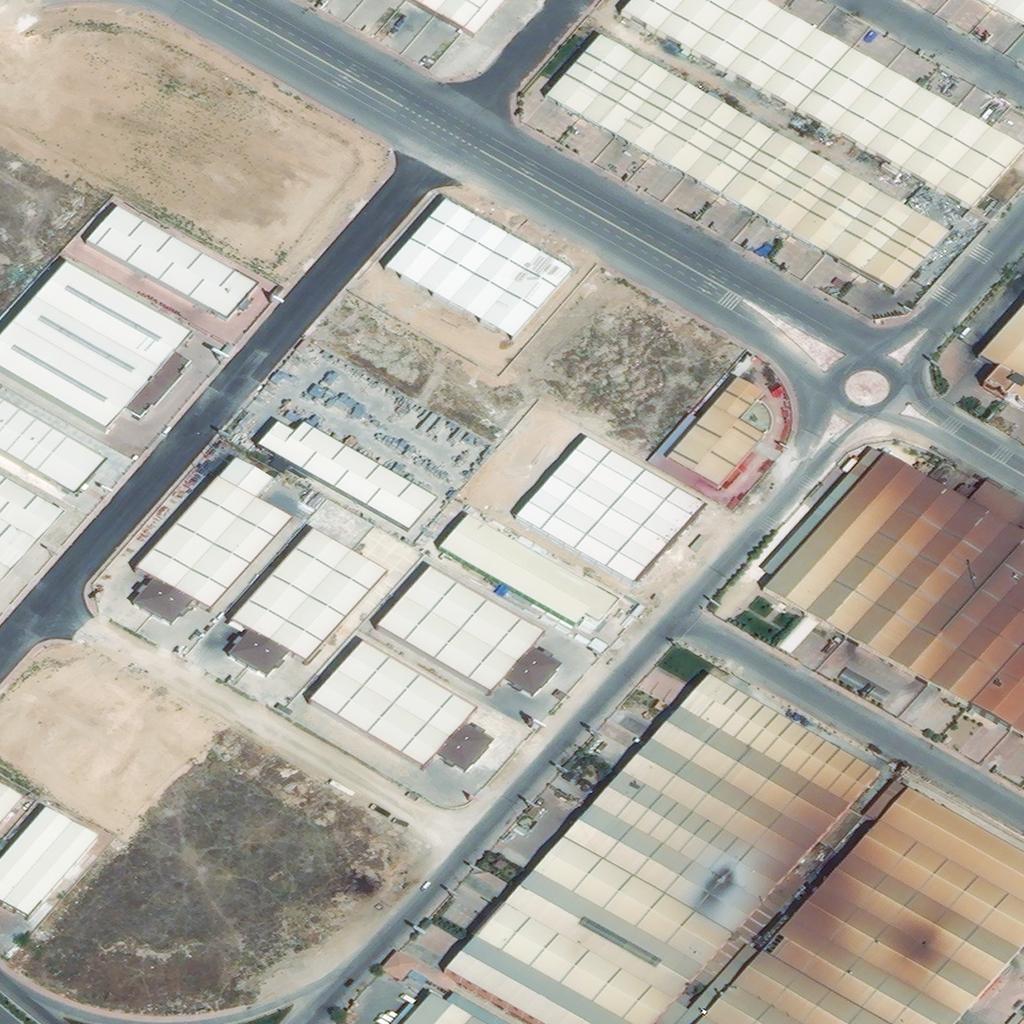} &
\includegraphics[width=0.24\linewidth]{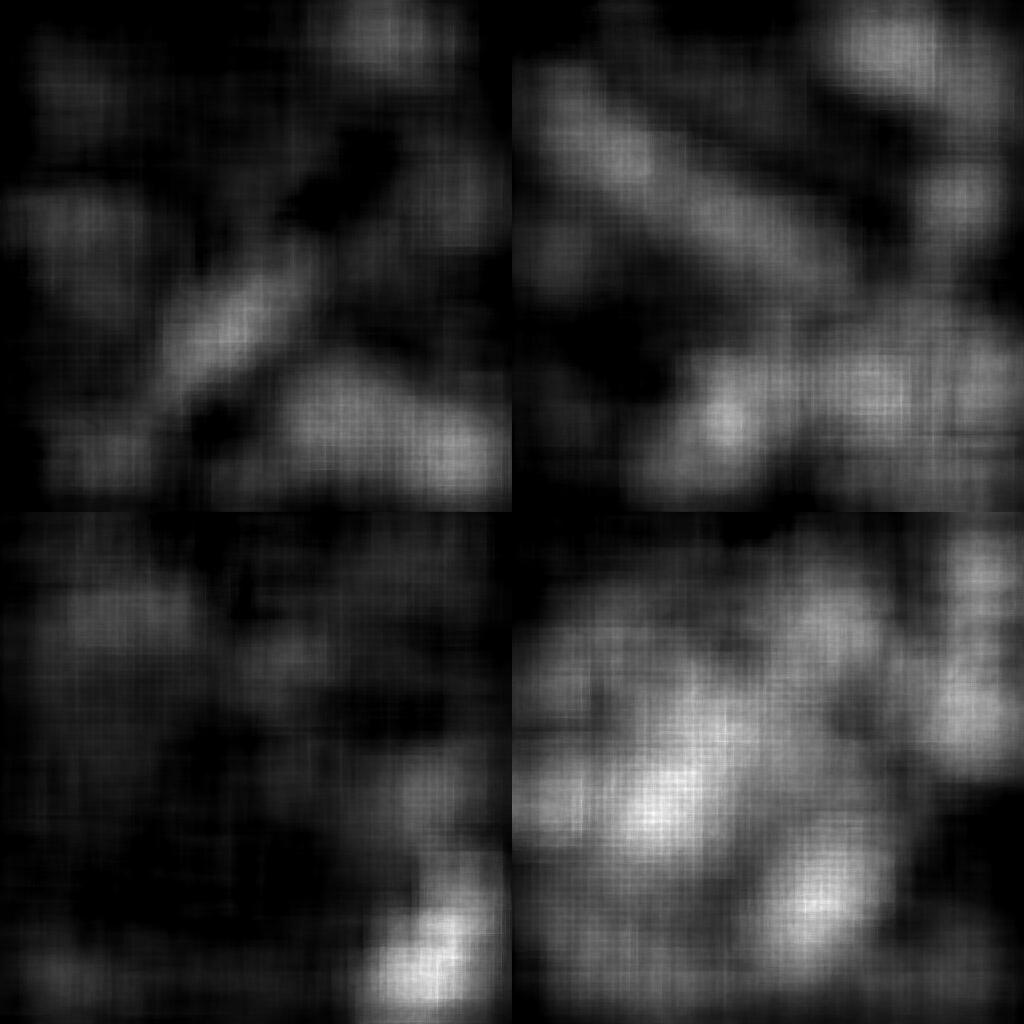} &
\includegraphics[width=0.24\linewidth]{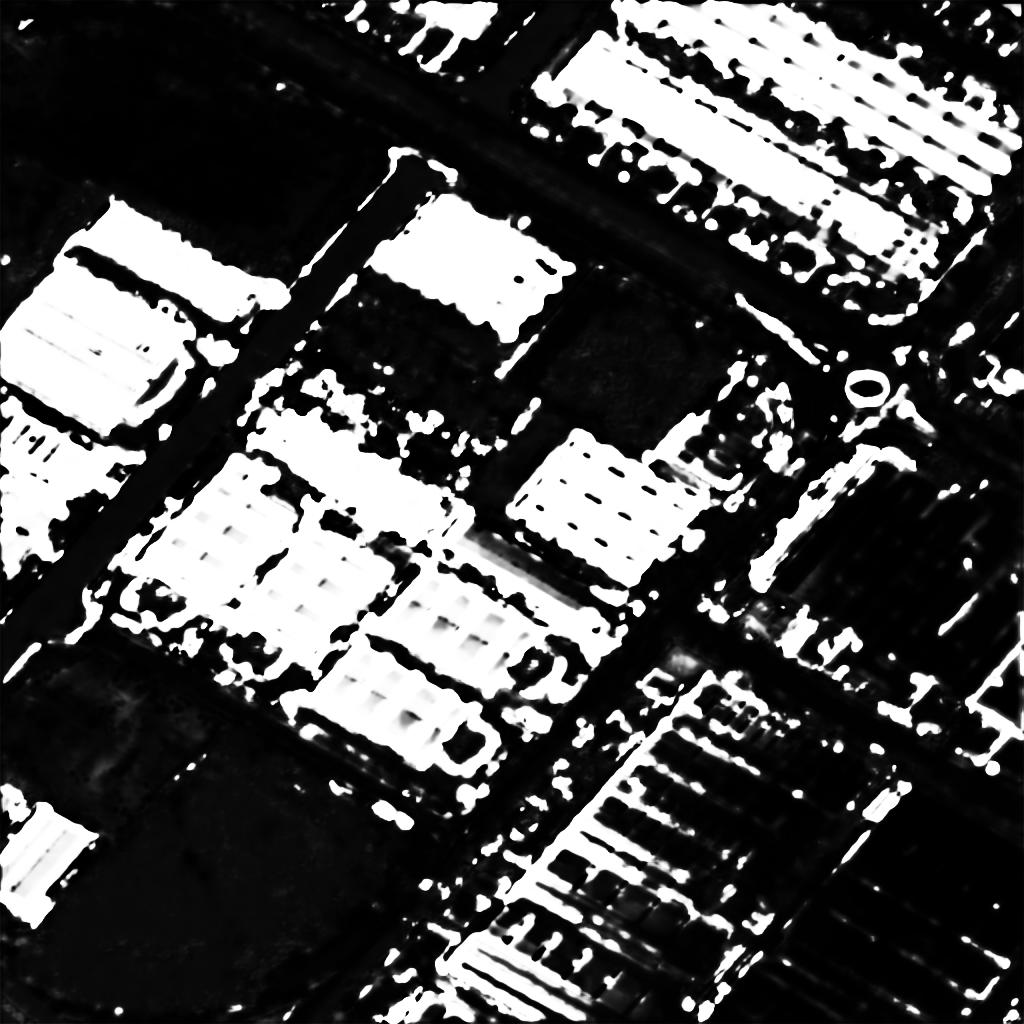} &
\includegraphics[width=0.24\linewidth]{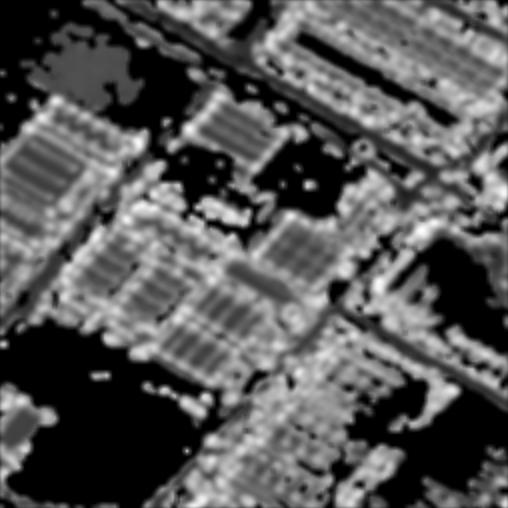} \\
\includegraphics[width=0.24\linewidth]{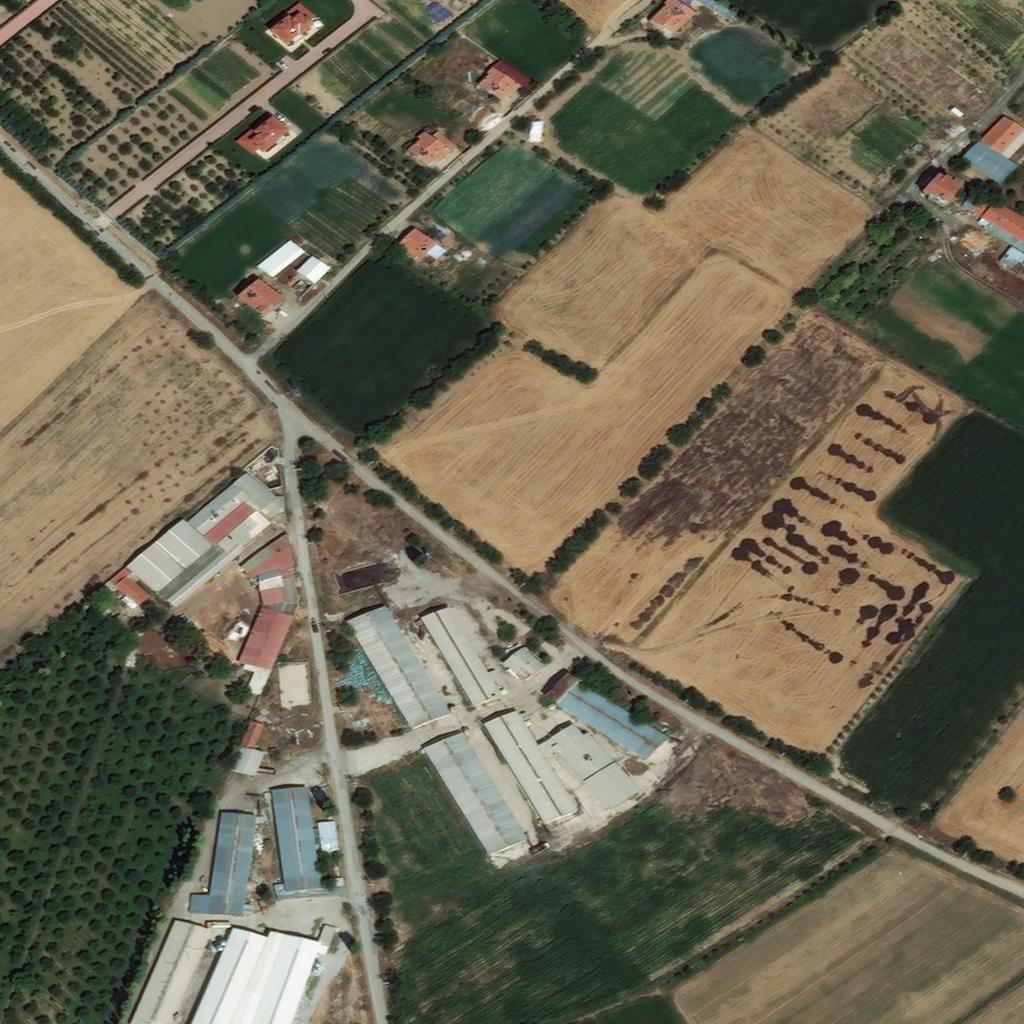} &
\includegraphics[width=0.24\linewidth]{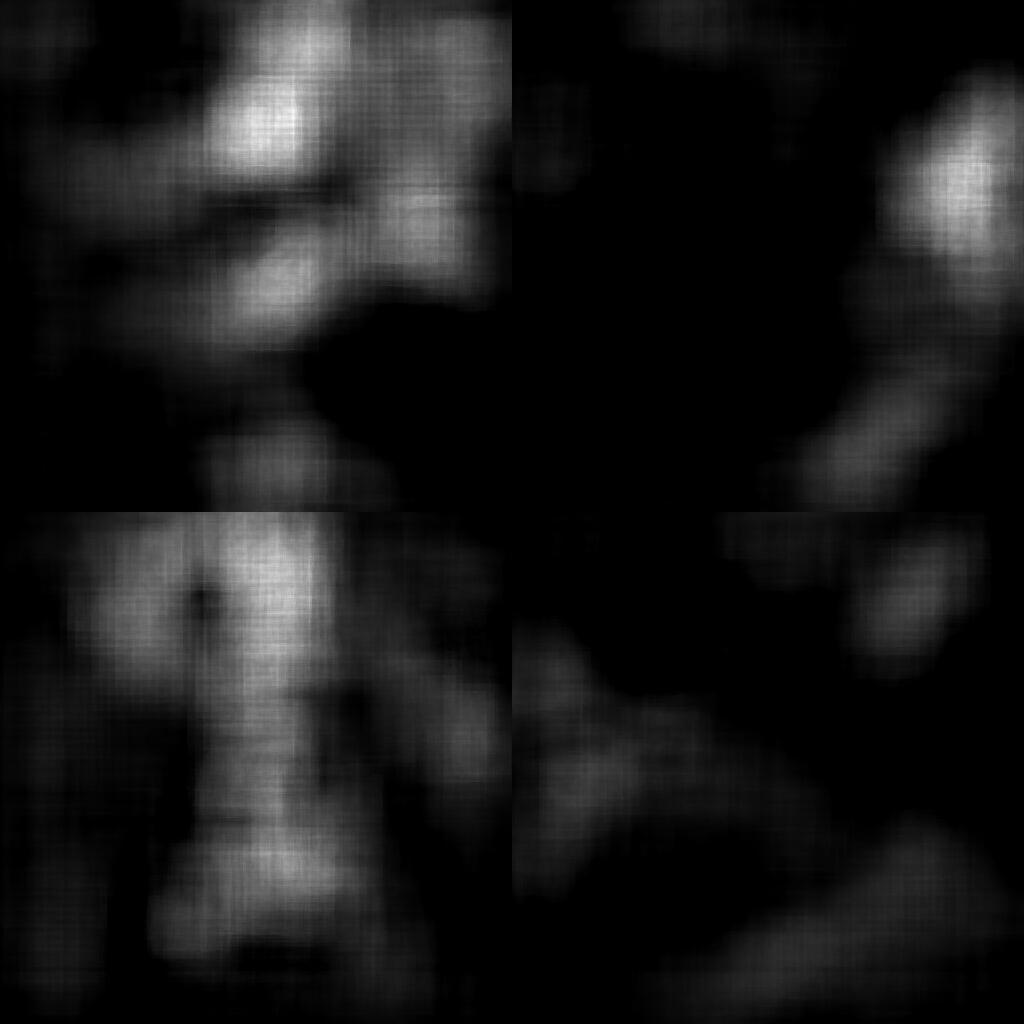} &
\includegraphics[width=0.24\linewidth]{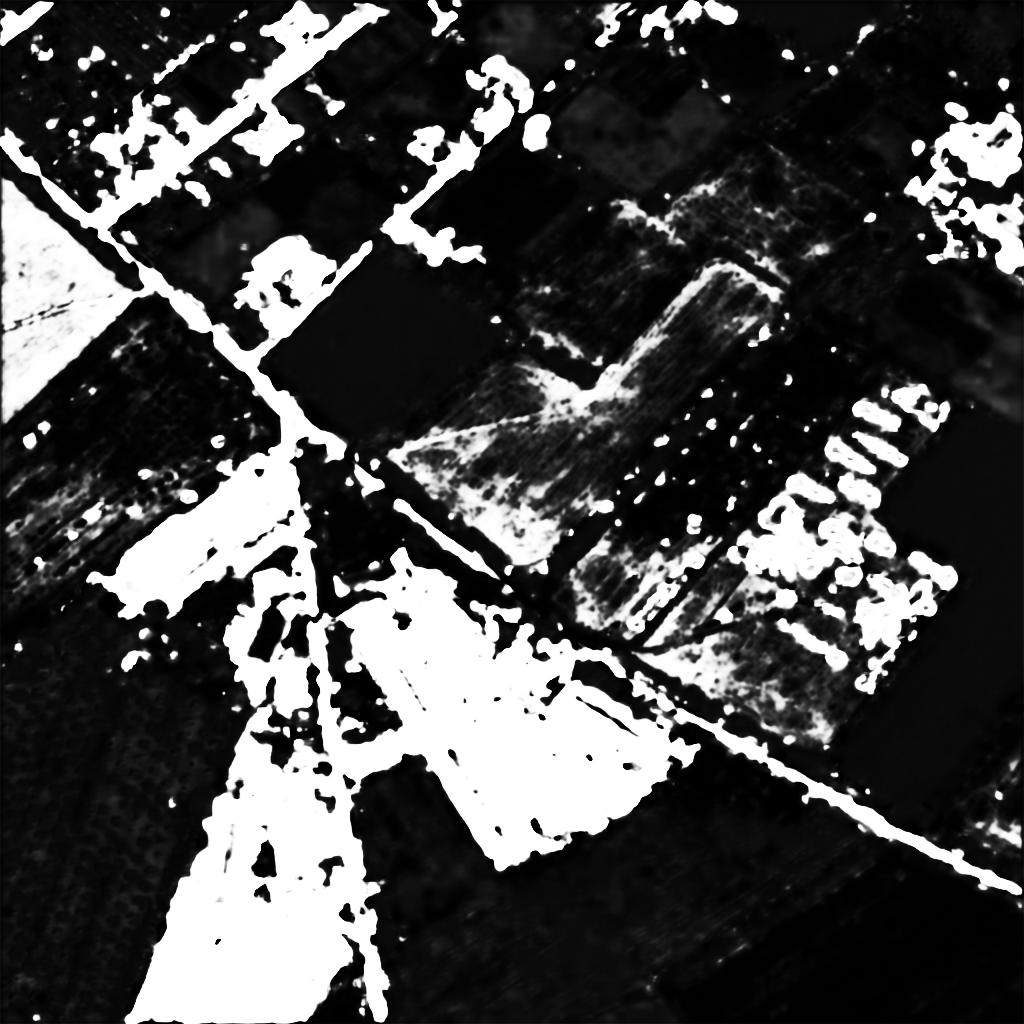} &
\includegraphics[width=0.24\linewidth]{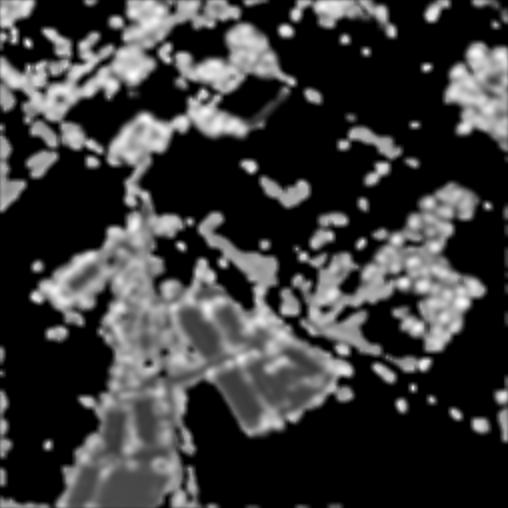} \\
\end{tabular}
\caption{Visualization comparison between the proposed method with U-Net \cite{ronneberger15_u_net} and SegNet \cite{badrinarayanan15:_segnet}. 
(Satellite images are from Digital Globe.)
}
\label{fig:visualization_comp}
\end{center}
\vspace{-10pt}
\end{figure}

\noindent
\textbf{Visualization Comparison}: Figure \ref{fig:visualization_comp} shows the visual comparison of the proposed model and all baseline models. We take the whole tile as input and compute the pixel-wise probabilistic maps as output to visualize.
As shown in the comparison, SegNet \cite{badrinarayanan15:_segnet} fails in almost all examples: it can not localize the object of interests in accurate position, and the probabilistic maps show little discriminative information.
U-Net \cite{ronneberger15_u_net} produces better localization results, but it is unable to localize objects with small size.
For our proposed method in this paper, we show the probabilistic maps before and after performing feedback. According to the comparison, the feedback strategy suppresses irrelevant pixels significantly. Moreover, it successfully localizes / segments most objects of interests with various spatial size.

\setlength{\tabcolsep}{0.4pt}

\noindent
\textbf{Quantitative Evaluation}: In Table \ref{tab:performance} we show the comparison of the proposed method with baseline methods, evaluated using F-Score.
From this comparison, the proposed method achieve significant improvement over all baselines.
Especially, from the comparison between before / after feedback in the network, we can see the feedback algorithm proposed in Algorithm \ref{alg:weakly-supervised} successfully suppresses a certain amount of irrelevant neuron activations.

\subsection{Discussion}

We summarize results from both the visualization and quantitative comparisons above. Firstly, the proposed algorithm can successfully learn from weakly-supervised data and produce high quality semantic segmentation results on our challenging data. Secondly, the feedback operation suppresses irrelevant neuron activations significantly as suggested from the comparison in Figure \ref{fig:visualization_comp} and Table \ref{tab:performance}. Also compared with other baseline methods, it achieves superior quantitative performance.

Another advantage is its efficiency, elegance and conciseness. The proposed method learns / predicts in an end-to-end fashion, without post-processing. This helps the network to short its inference time within seconds. Moreover, the Feedback Unit could also be applied to other network structures, and can be used in a variety of other applications such as visual-based auto-pilot and anomaly detection.

%% file: 6_conclusion.tex
\section{Conclusion}

In this paper, we propose a novel weakly-supervised neural network via Feedback strategy, to perform the task of image semantic segmentation. It suppresses irrelevant neuron activations in order to localize the object of interest.
An efficient algorithm is proposed to solve the optimization in a layer-wise fashion.
The proposed framework does not require additional post or pre-processing and is learned end-to-end.
We test this algorithm on the real-world application of performing cartography of human-made settlements derived from satellite imagery. It contains millions of training images without fully supervised labels and is evaluated on Terabytes of satellite imagery spanning multiple continents. Compared with several baseline methods, the proposed algorithm achieves superior performance, from both visualization and quantitative evaluation.

\noindent
\textbf{Future Work}
There is a tradeoff between footprint segmentation and classification performance. Models evaluated only on classification do poorly on semantic segmentation, and vice versa.  This is likely because models are concentrating on specific salient features, such as the corners of buildings, rather than learning to fill in the entire building contour.  A possible superior solution is to have two separate models each optimized for classification and segmentation respectively and combine them for better results.